\documentclass[runningheads]{llncs}
\usepackage{txfonts}
\usepackage{graphicx}
\usepackage{xspace}
\usepackage{relsize}
\usepackage{xcolor}
\usepackage{soul}
\usepackage{enumitem}
\usepackage{textcomp}
\usepackage[hyphens]{url}

\definecolor{cornflowerblue}{rgb}{0.39, 0.58, 0.93}
\definecolor{myGreen}{rgb}{0.094, 0.569, 0.22}

\newcommand{\webprotege}{Web\-Prot\'eg\'e\xspace}
\newcommand{\protege}{Prot\'eg\'e\xspace}
\newcommand{\rdfsSubClassOf}{{\smaller{\texttt{rdfs:subClassOf}}}\xspace}
\newcommand{\pinterest}{Pinterest\xspace}
\newcommand{\pin}{Pin\xspace}
\newcommand{\pins}{Pins\xspace}
\newcommand{\pinner}{Pinner\xspace}
\newcommand{\pinners}{Pinners\xspace}
\newcommand{\taxonomy}{Pinterest Taxonomy\xspace}
\newcommand{\tastegraph}{Pinterest Taste Graph\xspace}
\newcommand{\term}[1]{{\smaller{\textsf{#1}}}\xspace}
\newcommand{\ui}[1]{{\smaller{\textsf{#1}}}\xspace}
\newcommand{\attribute}{{\smaller{\texttt{Attribute}}}\xspace}
\newcommand{\attributes}{{\smaller{\texttt{Attributes}}}\xspace}
\newcommand{\interest}{{interest}\xspace}
\newcommand{\interests}{{interests}\xspace}

\begin{document}

\title{Use of OWL and Semantic Web Technologies\\ at \pinterest}


\author{\thanks{Authors are listed alphabetically}Rafael S. Gon\c{c}alves\inst{1} \and
Matthew Horridge\inst{1} \and
Rui Li\inst{2} \and
Yu Liu\inst{2} \and
Mark A. Musen\inst{1} \and
Csongor I. Nyulas\inst{1} \and
Evelyn Obamos\inst{2} \and
Dhananjay Shrouty\inst{2} \and
David Temple\inst{2}
}

\authorrunning{R. Gon\c{c}alves, M. Horridge, R. Li et al.}

\institute{Center for Biomedical Informatics Research, Stanford University, CA, USA \and
\pinterest, San Francisco, CA, USA}

\maketitle

\begin{abstract}
\pinterest is a popular Web application that has over 250 million active users.  It is a visual discovery engine for finding ideas for recipes, fashion, weddings, home decoration, and much more.  In the last year, the company adopted Semantic Web technologies to create a knowledge graph that aims to represent the vast amount of content and users on \pinterest, to help both content recommendation and ads targeting. In this paper, we present the engineering of an OWL ontology---the \taxonomy---that forms the core of \pinterest's knowledge graph, the \tastegraph.  We describe modeling choices and enhancements to \webprotege that we used for the creation of the ontology. In two months, eight \pinterest engineers, without prior experience of OWL and \webprotege, revamped an existing taxonomy of noisy terms into an OWL ontology. We share our experience and present the key aspects of our work that we believe will be useful for others working in this area.


\end{abstract}


\section{Introduction}
\label{sec:intro}

\begin{figure}[t]
    \centering
    \includegraphics[width=0.65\textwidth]{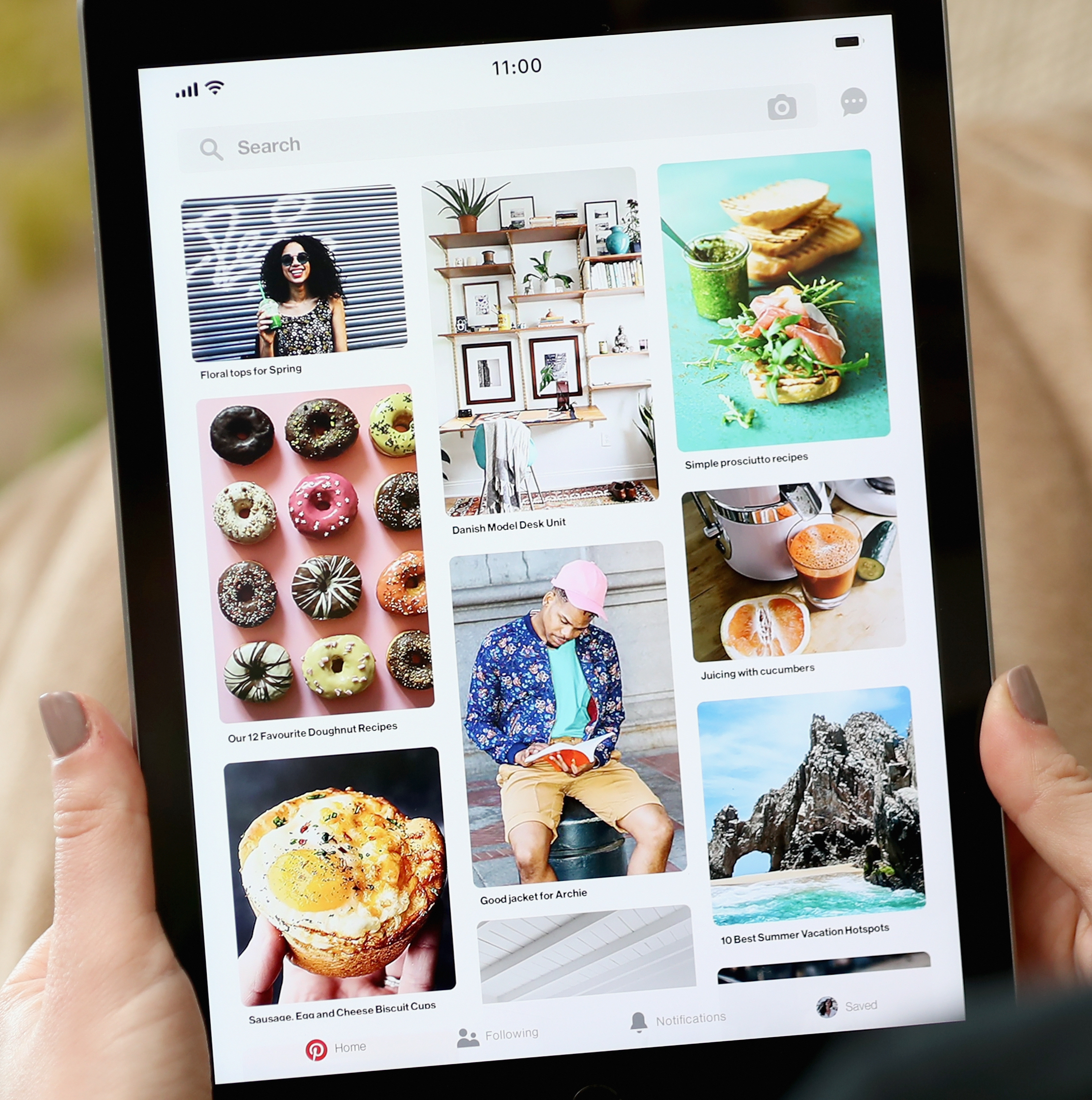}
    \caption{\textbf{An example of a \pinner's home feed.}  The feed displays examples of \pins that the \pinner has saved to their boards and suggested \pins that they might also be interested in. \pins are comprised of a representative image, a title and a snippet of text.  Clicking on a \pin will take the user to the Web content that the \pin represents or bookmarks.}
    \label{fig:pinterst}
\end{figure}

\pinterest \footnote{\url{https://www.pinterest.com}} was founded in 2010, and is headquartered in San Francisco, California. \pinterest offers a visual discovery engine that helps people find things that they like, which might be things they would like to do, such as scuba diving, places that they might like to visit, such as tropical islands, garments that they might like to wear, such as Bohemia dress, and so on. More specifically, \pinterest offers users a collection of digital pin-boards, or simply boards (Figure \ref{fig:pinterst}).  Users, known as ``\pinners'', save bookmarks for Web content, known as \emph{\pins}, to boards.  A \pin can be shared amongst boards and is visualized by an image that summarizes what the \pin represents.  Clicking a \pin takes a user to the underlying Web page that hosts the image and related content. 

Both \pins and \pinners are highly diverse and the amount of content is substantial---\pinterest hosts over 175 billion \pins and it has over 250 million monthly active users. To recommend the most relevant \pins to its users, and to achieve precise ads targeting, \pinterest defines a set of \interests. These \interests are simply terms that describe what each \pin/image on \pinterest is about, and what each user on \pinterest is interested in. The \interests are organized in a hierarchical structure, called ``the \taxonomy''. Behind the scenes, \pinterest categorizes both \pins and \pinners into one or multiple \interests. By knowing what a user is interested in and what each \pin is about in the same categorization space, it becomes easier to provide personalized recommendations. Advertisers can also use the \taxonomy to create Ads campaigns on \pinterest by selecting \interests from the taxonomy. The selected \interests essentially identify groups of \pinterest users who will be targeted by the campaigns.

In this paper, we describe the engineering process behind the \taxonomy, and we discuss key aspects of the work carried out by the \pinterest and \protege teams that we believe are relevant and useful to others working in this area. We discuss the use of OWL to model the content in \pinterest, and the benefits of using OWL over previous spreadsheet-based representations. We describe the \webprotege collaborative editing environment that was used to create, maintain, and evolve the \taxonomy, and we document the extensions to \webprotege that we implemented to optimize the \pinterest taxonomy construction workflow at \pinterest.

\section{Nomenclature}
\label{sec:preliminaries}

\begin{figure}[t]
    \centering
    \includegraphics[width=0.7\textwidth]{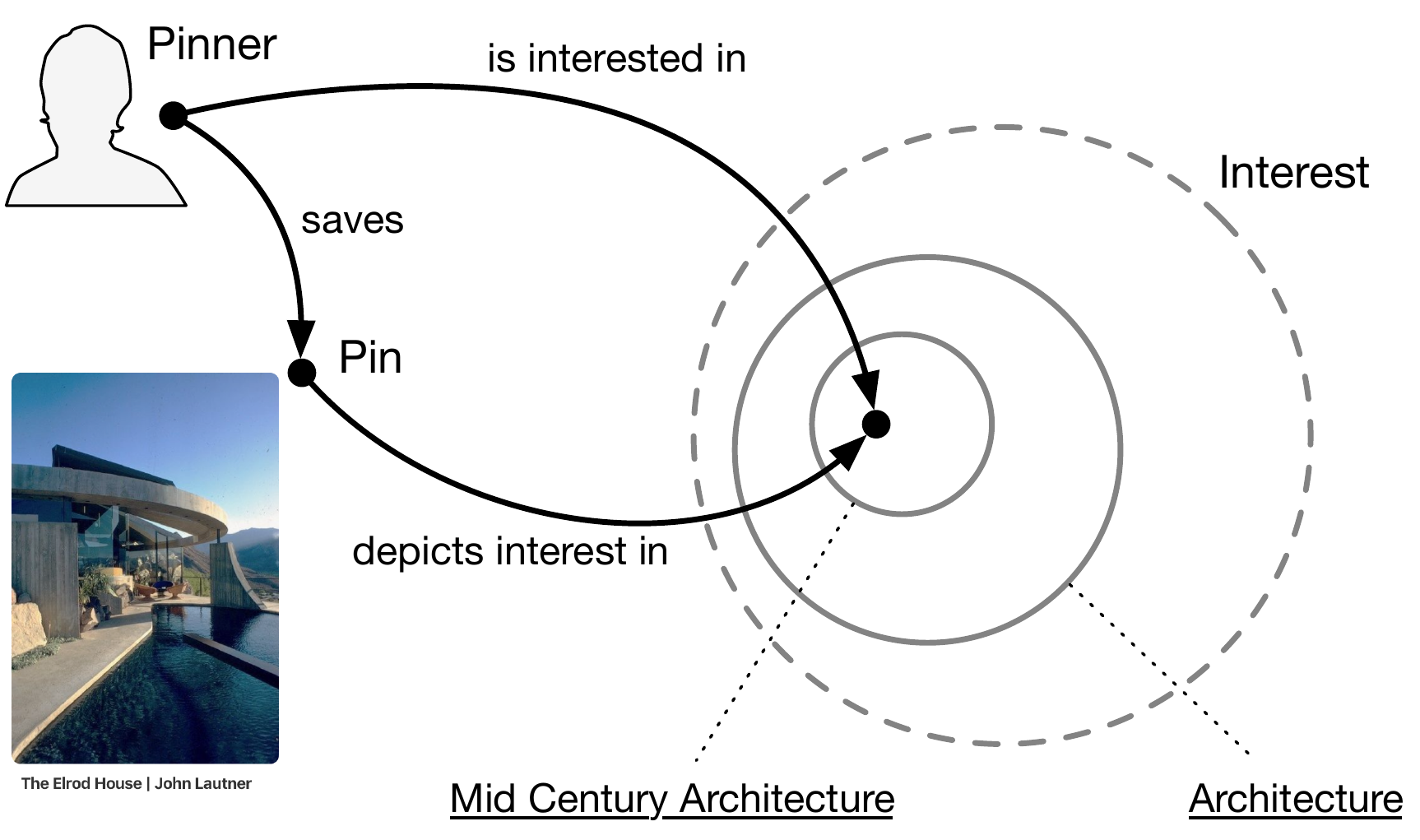}
    \caption{\textbf{A conceptual view of the \tastegraph}.  A \emph{\pinner} saves a \emph{\pin} to one of their \emph{Board}s.$^\star$  The \pin depicts/represents one of the \pinner's \emph{\interest}s.  Here, the \pinner is interested in ``Mid Century Architecture'', which is an interest in ``Architecture''.  Architecture is a top-level \interest, which is known as a \emph{Vertical}.  The hierarchy of \interests is known as the \taxonomy, or ``the taxonomy'' for short. ($^\star$\footnotesize{Note that the relationship of the \pin to the board is not shown here.})}
    \label{fig:tg-schematic}
\end{figure}

In what follows we define the nomenclature that we use throughout the rest of the paper.  Figure \ref{fig:tg-schematic} shows a schematic of the \pinterest nomenclature and represents an abstract view of the ``\tastegraph''.

\begin{description} 
    \item [\pin] An image, or visual bookmark on \pinterest that includes a description and links to an external URL.
    \item [\pinner] A \pinterest user who creates and/or saves \pins to their boards.
    \item [Interest] A concept that denotes what a \pin is about, or what a user is interested in. The Interest can be simply an answer to the question ``What are you interested in?'', which could be ``Photography'', ``Cooking'', etc.  Interests can be very broad, for example, ``Event Planning'' or ``Food and Drink'', or  specific, for example, ``Scuba Diving'', or very specific, for example, ``DIY Pom Pom''.
    \item [Board] A collection where \pinners organize their \pins in context as they plan. For example, a \pinner could create a board called \textit{Italian Recipes} and add ``Pizza Recipe'', ``Home Made Pesto Sauce'' and ``Veggie Lasagna'' \pins to it. 
    \item [Taxonomy] Extended from the scientific definition \cite{taxonomy}, the \taxonomy is a hierarchical arrangement of \interests. In OWL, the taxonomy roughly corresponds to the class hierarchy of an ontology.
    \item [Vertical] A top level node in the \taxonomy and its sub-trees.  Examples of verticals are ``Women's Fashion'', ``Home Decor'' and ``Architecture''.  Figure \ref{fig:tg-schematic} depicts a (very small) portion of the ``Architecture'' vertical.
    \item [Taste Graph] The \pinterest knowledge graph.  The Taste Graph~\cite{tastegraph} is a graph formed by combining the \taxonomy with nodes representing \pinners and \pins.  Every \pinner node and every \pin node is associated with one or more \taxonomy nodes. 
    \item [\webprotege] A collaborative cloud-based OWL ontology development environment.  A \webprotege user can log in to \webprotege, create an ontology project, and then share the project with geographically distributed collaborators.  Users see and discuss ontology changes in real-time.
\end{description}

\section{From Pins, to Interests, to a Taxonomy, to an Ontology}

\pinterest has come a long way in terms of understanding the \pins and users, and uses those to power its business. The company started out by keyword based understanding: extracting keywords from each \pin, doing clean ups, canonicalization, and labeling the \pins with these keywords. After that, based on users' engagement with the \pins, the users were labeled with some keywords too. Then \pinterest leveraged both the labels on the \pins and on the users to do recommendation and ads targeting. 

Two years ago, \pinterest decided to enable an ``interest'' based ads targeting interface, which named the most popular keywords as \interests, and organized them in a tree structure (a taxonomy) for advertisers to pick nodes from for their campaigns. For example, if Home Depot creates a campaign for the \interest ``Living Room'', both the users interested in ``Living Room'' in general, as well as users interested in ``Sofas'', ``TV Stands'', ``French Living Room Style'', ``Living Room Decor'', etc., will all be exposed to the ads of this campaign. Having realized the potential that such a structured knowledge representation provides, \pinterest decided to investigate ways to (1) improve the quality of the taxonomy; (2) augment the scope of the taxonomy as needed to provide coverage for all \pinterest content; and (3) settle on principled and robust processes by which the taxonomy is constructed, maintained, and reviewed.

Given the wide diversity and the (sometimes highly) specific nature of the \pins hosted on \pinterest, existing public taxonomies such as the Google Product Taxonomy\footnote{ \url{https://www.google.com/basepages/producttype/taxonomy.en-AU.txt}} are insufficient to fully describe \pinterest content. Thus, \pinterest decided to develop their own taxonomy, targeted at the kind of content that it hosts, and focusing on a specific business use-case: ads targeting. In the first version of the \taxonomy, \pinterest had manually defined and organized $\sim$400 \interests in a 2 level hierarchy, most of which were very broad. Subsequently, \pinterest released a far more granular version, with $\sim$6,000 \interests organized in a 3 level hierarchy. This version was entirely curated in a spreadsheet that was generated based on terms from top queries in \pinterest. One year later, based on feedback from advertisers, \pinterest decided to improve the taxonomy in terms of the quality and quantity of the interests in it. At that point, the \pinterest and \protege teams begun working on an OWL version of the \taxonomy.

\pinterest started out editing the \taxonomy in spreadsheets. Soon after, the \pinterest Content team realized that it was difficult to visualize, keep track of changes, and associate \interests with metadata. Thus, \pinterest decided to adopt a standard knowledge representation language and more appropriate tooling for collaborative taxonomy editing. \pinterest chose to use the Web Ontology Language (OWL) \cite{grau2008owl} to model their taxonomy, and \webprotege \cite{horridge2014webprotege} as the collaborative development environment. 

Since the adoption of OWL and \webprotege, \pinterest has greatly shortened the development cycle of the \taxonomy. The new tooling has made it possible for \pinterest to build an end-to-end system in just two months. In this time span, \pinterest designed an OWL ontology in \webprotege; built guidelines and workflows for human curation using \webprotege; loaded the 6,000 \interest taxonomy into \webprotege; enriched it with 5,000 new \interests extracted from user provided content; cleaned it up and re-organized it; and developed an engineering pipeline to consume the ontology, and use its content to populate a relational database for internal product consumption.

\section{Key Requirements}
\label{sec:requirements}

The insights that \pinterest drew from building the initial versions of the \taxonomy, and from advertisers' feedback, have provided us with concrete business requirements for this project. We list and discuss these requirements below. Subsequently, we enumerate the tooling requirements needed to fulfil the business requirements.

\subsubsection{Business Requirements}  The \taxonomy will be used internally to categorize all the \pins and all the users, and externally to power \pinterest's ads targeting. To fulfill both of these use cases, the final knowledge representation needs to:

\begin{enumerate}
    \item Be a single root tree structure, instead of a directed acyclic graph (DAG) for near-term downstream consumption.  It should however be possible to evolve the taxonomy into a multi-parent DAG;
    \item Provide support for adding attributes (facets) to the \interests, in order to support multiple perspectives of the categorization and poly-axial classifications;
    \item Match the \pinterest content, that is, it should include \interests that depict a substantial number of \pins and exclude \interests that \pinterest has little or no \pins for;
    \item Contain no ambiguous \interests ---the \interests' names should be clear on their own even after removing the context provided by the tree structure (i.e., the parents in the hierarchy), for example, ``cricket'' the insect versus the sport;
    \item All children of the same parent should be mutually exclusive and collectively exhaustive (MECE); and
    \item Quality of the \interests is more important than quantity.
\end{enumerate}
 
\subsubsection{Tooling Requirements}  To manage, curate and evolve the \taxonomy, \pinterest needs a tool that provides:

\begin{enumerate} 
    \item The ability for multiple editors located in different geographical locations to work on the same project simultaneously;
    \item A way to track which \interests in the taxonomy have been reviewed;
    \item The ability to efficiently reorganize the taxonomy---move an \interest to a different branch, merge, rename or deprecate it;
    \item A way to add annotations/metadata to one or multiple \interests, e,g., sample \pins, a short description of the \interest, synonyms, statistics and attributes of this \interest;
    \item Multi-lingual support, that is, support for adding labels in multiple languages and for displaying the full taxonomy in different languages; and
    \item A friendly user interface (UI) to allow people to browse and search for \interests in the taxonomy based on their annotations. The UI should provide the ability to share links directly to content in the taxonomy.
\end{enumerate}

\section{Ontology Modeling Experiments}
\label{sec:taxonomy}

We conducted several ontology modeling experiments.  The goals of these experiments were:
(1) To determine the kind of vocabulary defined in Section \ref{sec:preliminaries} that we would need for the project; (2) To settle on (best practice) conventions, for example, rules for consistent naming of the \interests; (3) To experiment with editing workflows and define the curation instructions; (4) To evaluate \webprotege as a tool to satisfy editing requirements; and (5) To see what gaps needed to be filled in terms of tooling. 

\subsection{Deriving a Seed Ontology}
\label{sec:seed_terms}

We used the 6,000 \interests from the three-level taxonomy to bootstrap a starting ontology. There were several problems with the ontology output from this process:

\begin{description}
    \item [Lack of Coverage]  The \interests were not enough to describe all of \pinterest's content.  For example, it had 120 \interests for ``Men's Fashion'', over 400 \interests for ``Women's Fashion'', but no \interests at all for ``Children's Fashion''.
    \item [Imbalanced Structure] The taxonomy was very broad and shallow.  Moreover, it was inappropriately imbalanced with respect to the number of children per parent.  For example, one vertical had only 2 child \interests while another had over 80.  This may feel odd to advertisers and may be harder to find the relevant interests.
    \item [Irregular Precision] Some areas of the taxonomy were too fine-grained.  For example, the part of the taxonomy representing the ``Art'' vertical contained many \interests of the form ``11 x 17 posters'' or ``36 x 48 posters'' (representing \interests in very specific poster sizes).
    \item [Irregular Naming] The naming convention for the terms in the taxonomy was not uniform.  Some terms were named in singular form while their siblings were named in plural form.  There was also an inconsistent use of prefixes and suffixes.
\end{description}

\subsection{Detailed Modeling Pilot Study}

Having identified initial problems with the seed ontology, we honed in on two verticals, ``Home Decor'' and ``Fashion'', in order to focus on more detailed modeling issues and to get a better feeling for development environment issues.  We chose these verticals for the richness and variety of \interests contained in them (to expose modeling issues) and for their prominence on \pinterest.  We used seed lists of \interests to start constructing ontologies representing these verticals.

\subsubsection{Development Tools}
\label{sec:dev_workflow}

During the pilot study phase, we used a number of tools for engineering and communication that we describe next.

\paragraph{Collaborative Editing Environment} We used \webprotege and its collaboration features throughout the pilot experiment.  We made use of threaded discussions to document editorial decisions and point to external references that were considered.  We made heavy use of email and Slack notifications, which enabled timely responses to discussion within \webprotege.  We used the change tracking feature of \webprotege to review recent changes when starting a modelling session, and we used the ``live project feed'' to monitor current activity.

\paragraph{Communication Tools} We used Slack for communication outside \webprotege. Both the \pinterest and \protege teams were already familiar with this tool for internal communication.  Because \webprotege supports ``deep linking'', it was easy to paste links to entities in \webprotege directly into Slack.  We used Slack for any discussions unrelated to the ontology content, for example, to set up meetings, to discuss tooling matters, and to report and discuss software issues. We also held teleconferences on a regular basis.

\paragraph{In-person Meetings} We met face-to-face for extended periods of time at the start of the project, in order to make major decisions on workflow and tooling.  We held a workshop meeting early in the pilot experiment to simultaneously work on the ``Fashion'' vertical in the same room.  This enabled us to quickly assess usability and tooling issues, and it also helped us to quickly identify and discuss broad modeling issues.

\subsubsection{Design Decisions and Modeling Choices}

The modeling pilot study enabled us to settle on various modeling choices and engineering conventions:

\paragraph{Interests as Classes} We decided to represent \interests as classes.  This may seem odd, but \emph{ontologically}, an instance of an \interest represents someone's (a \pinner's) own particular, unique interest in something.  Thus, classes in the ontology represent interests and not the actual subject of an interest.  From herein we use \interests and classes interchangeably.

Using classes for \interests also side-steps the thorny issue of classes versus individuals.  Suppose someone is interested in San Francisco.  Ontologically, San Francisco, the place, is an individual.  That is, there is just one San Francisco in the domain of discourse (the world).  In the taxonomy, we do not explicate the fact that an \emph{\interest in San Francisco} represents an interest about San Francisco the place.  While this may seem straightforward, the water is much more muddy when one thinks about things like recipes, computer games or cute videos of cats.  Thus, focusing purely on \interests as classes helps to keep the modelling clean and simple, and helps to avoid overly complex debates about modeling.

One more important benefit of modeling \interests as classes is that \interests can be easily specialized.  This includes obvious cases of specialization, such as an interest in mid-century architecture is an interest in architecture.  It also includes less obvious specializations, such as an interest in 1960's San Francisco is an interest in San Francisco.  

\paragraph{Interest Descriptions} For each \interest, we added a label (its preferred name), plus synonyms (if available) and definitions (where warranted). We recorded this information using the following annotation properties:
\begin{description}
    \item [\term{rdfs:label}] as the primary, preferred name/label for an \interest in a given language.\footnote{Note that we could have chosen \term{skos:prefLabel}. We ensure that all labels are unique, so \term{skos:prefLabel} annotations could easily be generated from \term{rdfs:label} values.} Every label has a language tag, for example, \term{@en}.
    \item [\term{skos:altLabel}] for recording synonyms. We encoded all known synonyms of each interest to support our search and presentation goals.
    \item [\term{skos:definition}] to include a 1-2 sentence textual definition to clarify the meaning of an \interest. This is important, as it provides a shared understanding among the team of what this \interest is, especially when it is not a well-known term. Many of the definitions were copied from Wikipedia.
    \item [Domain specific annotation properties] for both business usage and curation usage. For example, there are certain \interests that are sensitive or brand related, and thus cannot (according to \pinterest policy) be exposed to advertisers.  These \interests are marked as \term{noAds=true} in order to identify them in the engineering pipeline and avoid exposing them in the targeting interface. We also used defined properties such as \term{isHumanReviewed}, to indicate whether an \interest has been human reviewed or not, and to eliminate the possibility of curators reviewing previously reviewed \interests.
\end{description}

\paragraph{Naming Conventions} We used title case names, with spaces, for interest names for example, ``Garden Bench''.  Ontology engineering recommendations often state that singular noun forms should be used for entity names \cite{montiel2011style,noy2001ontology,owlpizzas,schober2007towards,svatek2010entity}.  However, we determined that it was necessary, and more natural, to use a mix of singular and plural forms based on the forms used in \pinterest's top queries.

We attempted to normalize the names of \interests in a principled way.  For example, under ``Home Decor Styles'' there are a large number of styles.  Many of these were not uniformly named.  Some were named ending with ``Style'' or ``Styles'' (for example, ``California Style''), others were named ending in ``Interior'' or ``Interiors'' (for example, ``Art Deco Interiors''), while others were named ending in ``Decor'' (for example, ``Bohemian Decor'').  In these cases, we settled on particular patterns, depending upon the context, and then normalized \interests according to these patterns.   Whenever we renamed a topic we endeavored to preserve the original name in a \term{skos:altLabel} annotation to keep the old name as a synonym.

\paragraph{Name Ambiguity} Some topic names in the original \pinterest taxonomy were used in different senses. For example, ``Topiary'' is both the activity of sculpting plants into three-dimensional shapes, and plants themselves that have been sculpted this way.  We disambiguated them in the way that thesauri entries are disambiguated, for example, ``Topiary (Plant)'' and ``Topiary (Gardening Activity)''.

\paragraph{Interest Disambiguation} We frequently had to use the \pinterest text-search functionality to disambiguate obscure names. We struggled with ambiguous names like ``privacy screen'', ``water scooper'', ``valances'', among others.  We made extensive use of Wikipedia for providing concise textual definitions for \interests to aid curators during the review process.  With an eye to advertising---one of the most important uses of the taxonomy---we also viewed external Web sites, such as Bed, Bath and Beyond, Crate \& Barrel, IKEA and Walmart in order to compare their product categorization with the kinds of products that are represented by \interests.

\section{Production Tooling}
\label{sec:tool_support}

Given the working relationship between \pinterest and Stanford, \webprotege was the obvious choice for an editing environment.  The pilot experiments that we carried out revealed that \webprotege was able to meet the majority of the tooling requirements.  However, we significantly enhanced \webprotege to satisfy previously unmet requirements and, in places, to streamline existing cumbersome editing operations.  In what follows, we describe the key aspects of \webprotege in the context of the tool requirements for this project.

\begin{figure}[t]
    \centering
    \includegraphics[width=\textwidth]{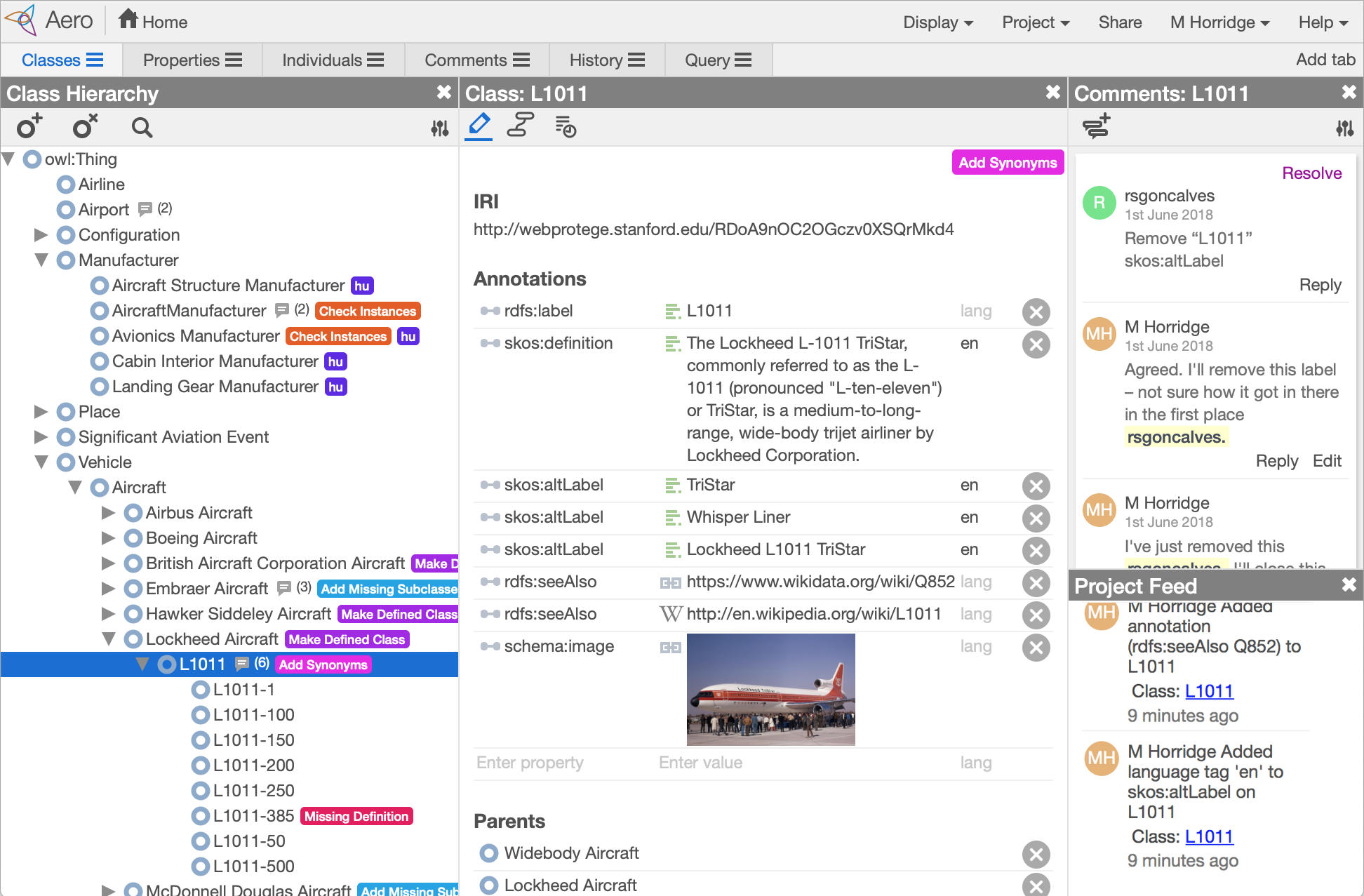}
    \caption[The \webprotege UI]{\textbf{The \webprotege UI.} The figure shows the main \webprotege user interface displaying the `Aero' project$^\star$.  The left hand side displays the class hierarchy, containing various ``tagged'' classes.  The center displays content for the selected class, in this case L1011. This center panel can be switched to display the history for the selected class.  The right hand side displays threaded discussions for the selected class and recent editing and commenting activity.  ({$^\star$}Note that, this project bears no relation to the \pinterest project.  It is used here merely for illustration purposes.)}
    \label{fig:wp-edit}
\end{figure}

\paragraph{Support for Multi-User Collaboration and Sharing}

\webprotege is a cloud-based OWL ontology development environment where users perform editing and viewing tasks in a Web browser.  It allows geographically dispersed users to collaborate in real-time.  

When a change is made to an ontology, all collaborators see the change in real time.  Changes are also tracked, in the form of axiom additions and removals. Information about a change is captured along with metadata about who performed the change and when the change was performed. The complete project change history is available for collaborators to peruse. \pinterest uses the change history to keep track of contributions by curators, and to carry out downstream analyses.

Crucially, \webprotege users can be assigned different roles within the context of a project.  This makes it possible to cater for workflows that comprise multiple teams with different responsibilities.  In the case of the \pinterest project, there is a core team of editors surrounded by a larger group of reviewers/commenters.  While the ``under the hood'' role management capabilities provided by \webprotege are very fine-grained, the default user-interface supports high level coarse-grained privileges, namely ``Manage'', ``Edit'', ``Comment'' and ``View''.  At the start of the project, it was not clear to us whether these privileges would be sufficient.  However, as editing and refinement of the ontology proceeded, it became clear that these basic permissions worked well enough.

\paragraph{Provision of a User Friendly Interface}

\webprotege provides editing support for the complete OWL 2 syntax.  However, by default, \webprotege displays a simplified editing interface (Figure \ref{fig:wp-edit}) that we believe is sufficient for the majority of ontology projects~\cite{HoTuVeNyMuNo13}.  

This interface allows users to edit ontologies in a frame based style, specifying parents (\rdfsSubClassOf under the hood), annotations, and relationships (\rdfsSubClassOf with super classes that match specific patterns of class expressions under the hood) in an intuitive frame-like way.  An image of this default user interface is displayed in Figure~\ref{fig:wp-edit}.  

So far, this simple interface has worked well for the \pinterest project, with the bulk of editing being annotation based editing and hierarchy editing.  \pinterest ontology curators required little training in how to use the interface---they attended a one hour training session on the \webprotege interface and ontology editing conventions.

\paragraph{Support for Ontology Reorganization}

As mentioned previously, the initial input for the \pinterest ontology was a spreadsheet that had been derived from user data.  This provided a seed taxonomy with at most three levels of depth and mixed bags of non-unique terms at each level.  One of the perceived benefits of moving to OWL and using \webprotege was that the ontology would be easier to browse and edit compared to the existing spreadsheet based approach.  While this has largely proven to be true, we had to extend \webprotege with two new features for streamlining the editing workflow.   

The first feature that we added was a workflow to \emph{merge entities}.  This allows multiple entities to be selected and then merged into a target entity.  This operation performs a number of complex steps under the hood, such as replacing references to the entities being merged with a reference to the target entity.  It leaves merged entity IRIs intact, but deprecates them and preserves annotations on them, for record keeping purposes.

The second feature that we added was a \emph{bulk move} operation.  The initial cleanup step involved a significant amount of re-organizing edits to be performed, sometimes between large disparate branches of the taxonomy hierarchy.  The new bulk move feature allows multiple entities to be selected and in the next step a new parent entity to be chosen for them.  While simple, this feature proved to be much more effective and more reliable than using drag and drop.

Finally, both merge operations and move operations typically involve multiple atomic changes to achieve the desired outcome.  \webprotege bundles up these atomic changes into a single composite change operation, which can then be applied with a manually entered commit message that appears in the change history log of the project.

\begin{figure}[t]
    \centering
    \includegraphics[width=\textwidth]{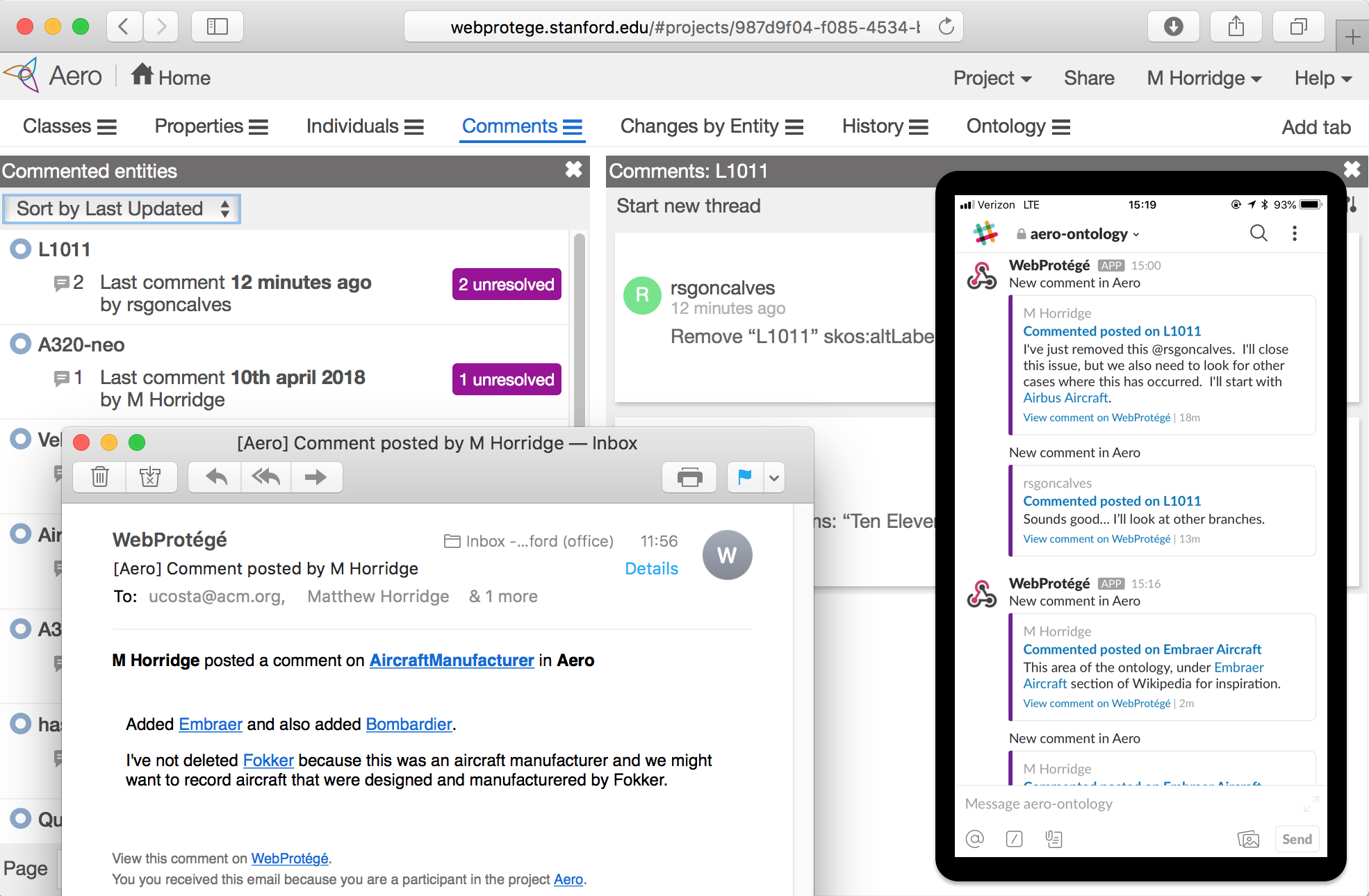}
    \caption{\textbf{Comments management and notifications.} The main part of the figure shows the \ui{Comments} tab. Comments can be sorted and viewed by entity. The lower left-hand side inset shows a notification email sent out to participants after a comment has been posted.  The right-hand side inset shows integration with the chat app Slack.}
    \label{fig:wp-comments}
\end{figure}

\paragraph{Support for Metadata Editing}

Interest classes in the ontology are richly described with entity annotations.  These annotations can be roughly split into two types: (1) \emph{Content description annotations}, which provide synonyms, descriptions, visibility flags, and pointers to example \pins; and (2) \emph{Status annotations}, which provide ``housekeeping'' information about the editorial status of classes.  In both cases we quickly learnt that there was a reoccurring desire to apply edits to a large number of annotations at once.  From switching a status flag from `false' to `true', to applying a consistent naming convention based on regular expressions.  We therefore added a powerful bulk annotation editing interface to \webprotege.  This feature allows entity annotations to be ``selected``, using patterns, and then modified, deleted or augmented based upon the criteria used to select them.

\paragraph{Support for Reviewing and Quality Control}

Besides offering real-time distributed editing capabilities, \webprotege provides support for collaborative interaction in the form of ``entity discussion threads'' (Figure \ref{fig:wp-edit}, right hand side).  Discussion threads can contain user mentions, links to entities and links to external resources.

This discussion functionality has been used for a number of purposes in the \pinterest project, including for filing \interest related issues and for soliciting reviews of \interest descriptions.  Issues were sorted and managed via the ``Comments'' tab (Figure \ref{fig:wp-comments}), which provides basic functionality for sorting issues by creation or modification time and by entity, and largely proved sufficient for the task at hand.

\begin{figure}[t]
    \centering
    \includegraphics[width=0.8\textwidth]{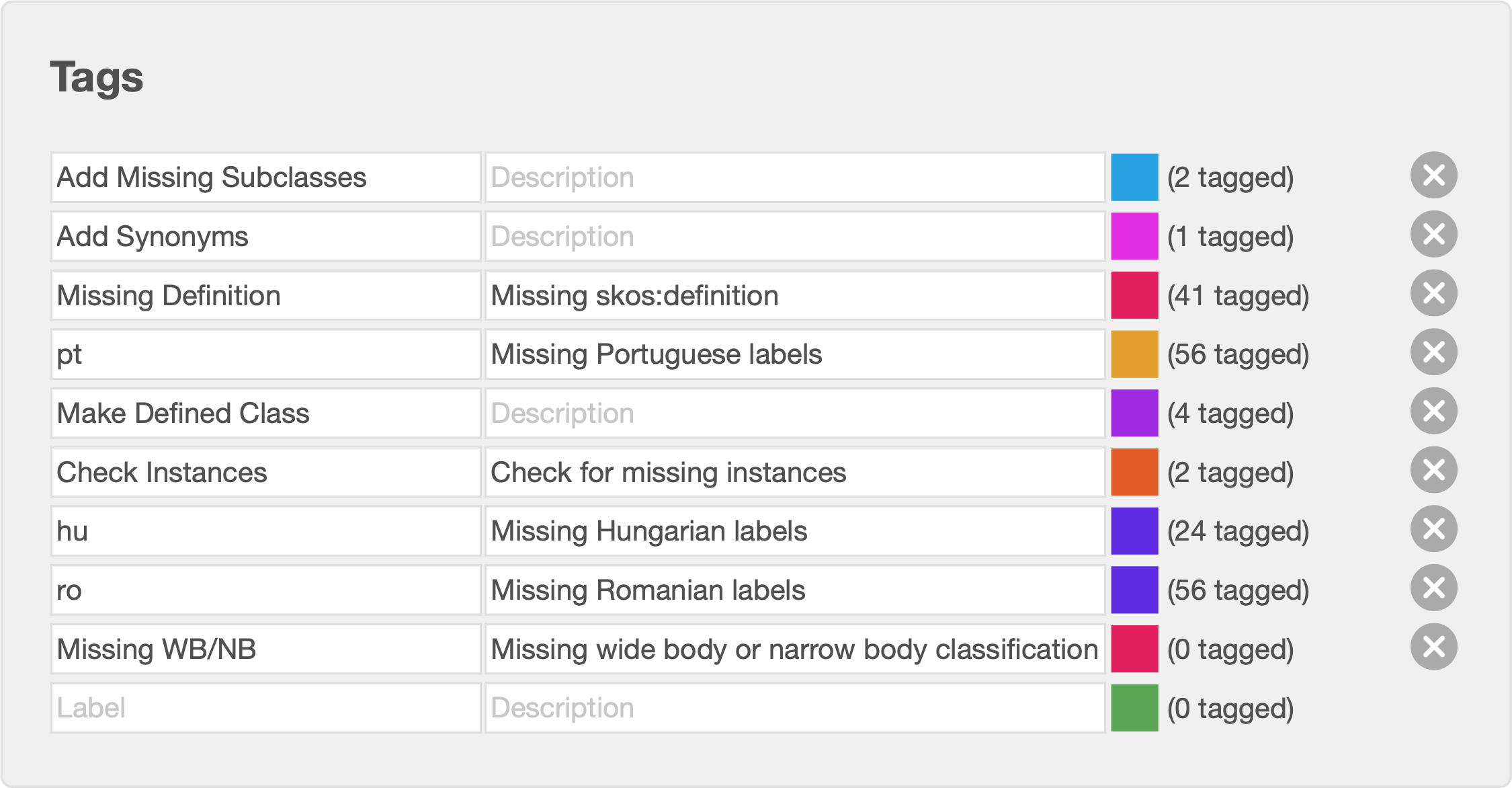}
    \caption{\textbf{Tags management in \webprotege.}  Each tag is assigned a label, an optional description and a color. Tags can be defined on a per project basis.  These tags can then either be manually assigned to entities or they can be assigned in an automated live query-based way.}
    \label{fig:tag-list}
\end{figure}

When discussions are posted to a project, collaborators are notified via email or Slack\footnote{\url{https://slack.com}} (Figure \ref{fig:wp-comments}) so that they do not miss discussion posts that are relevant to them.  Notification emails contain deep links to the \interests being discussed so that it is possible to ``jump'' straight to the relevant portion of the ontology in \webprotege.  This proved to be an effective way of engaging members of the \pinterest knowledge management team in active discussion.

As part of the taxonomy quality control process, all classes representing \interests require human review. Classes that require a review are flagged with annotations to indicate the review status.  Early on in the project, \pinterest requested custom highlighting for classes, so that certain classes, for example those requiring a review, would stand out from other classes.  To support this, we added ``entity tags''\footnote{Recall that entities, in OWL, are classes, properties, individuals and datatypes.} to \webprotege.

Entity tags can be used to highlight entities in a colorful, enticing way in the \webprotege user interface.  Example entity tags are shown on the left hand side of Figure~\ref{fig:wp-edit}, where there are tags for flagging missing definitions and missing Hungarian labels, among others.  Multiple tags can be specified for a given project, as shown in Figure \ref{fig:tag-list}, and multiple tags can be assigned to a single entity.  Not only do tags ``call out'' entities in the user interface, they are also searcheable.  Thus, it is possible to list and filter entities that have given tags.

\begin{figure}[t]
    \centering
    \includegraphics[width=0.8\textwidth]{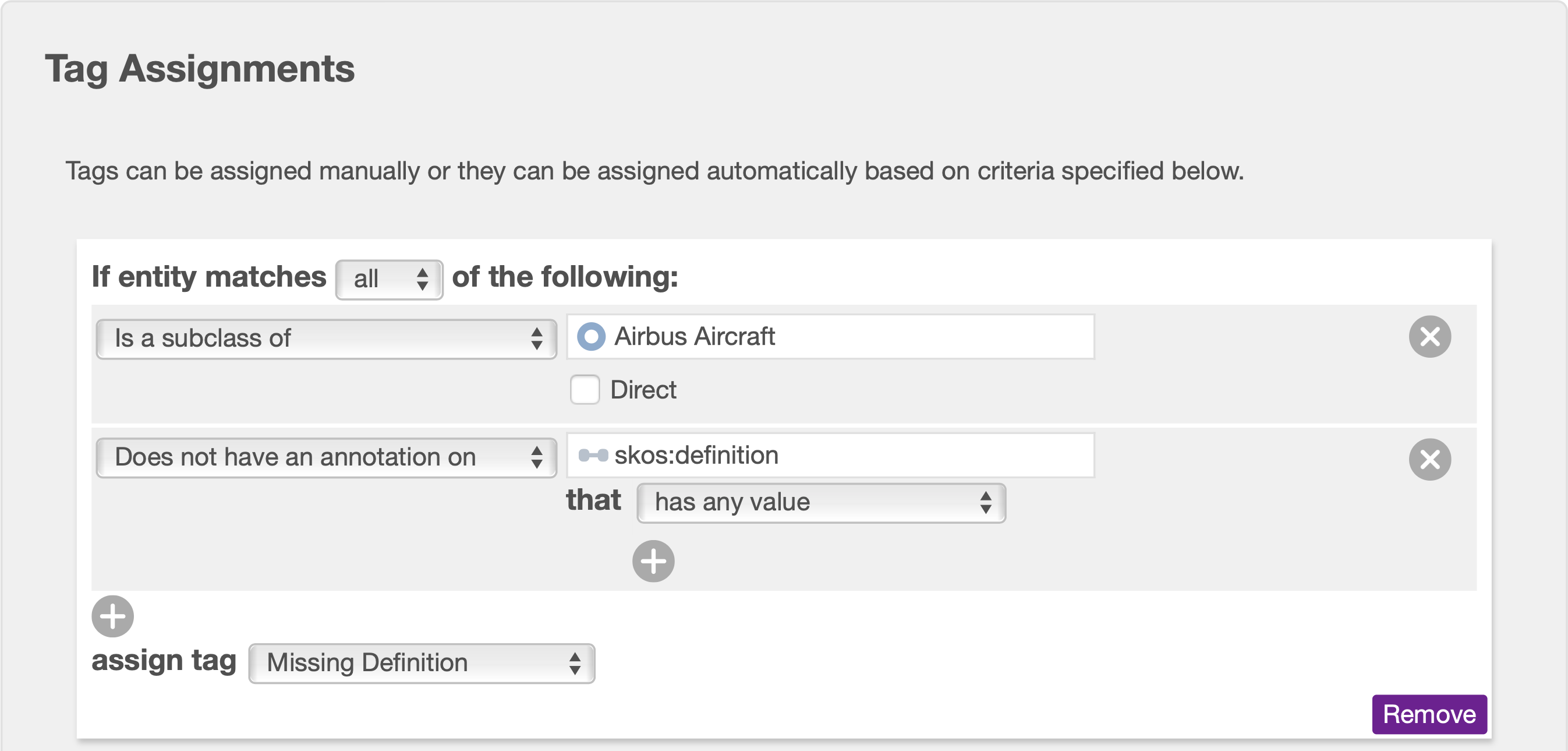}
    \caption{\textbf{Automated tag assignment.} Tags can be automatically assigned to entities based upon ontology structural criteria.  Here, the criteria specify that any descendants of ``Airbus Aircraft'' that are missing a value for \term{skos:definition} will be tagged with the ``Missing Definition'' tag.}
    \label{fig:tag-assignment}
\end{figure}

We designed Entity Tags so that they could either be assigned to entities in a manual, explicit fashion, or assigned in an automated manner based on ontology content.  Figure \ref{fig:tag-assignment} shows the set up page for automated tag assignment.  While a presentation of the full tagging capabilities is beyond the scope of this paper, it is possible to tag entities that match a given set of rules/criteria.  This tagging feature supports complex, multiple conjunctive and disjunctive criteria, along with paths of values to be matched.  Many types of matches are possible, such as matches by specific value, parts of values (regular expressions) and ranges.  Furthermore, it is possible to use entity matching criteria to check constraints that involve multiple values, such as label uniqueness (in the context of a given language) and annotation value disjointness (to enforce rules such as preferred labels being disjoint from alternative labels).

It is worth noting that the tagging functionality, that is, the entity matching criteria coupled with automated tag assignments, comes close to a SHACL Core \cite{knublauch2017shapes} implementation in a user-friendly guise.

\paragraph{Support for Multi-lingual Editing and Viewing}

\webprotege has always had support for specifying IETF language tags~\cite{bcp47} via auto-completion in the default editing interface.  However, it soon became clear that the \pinterest curators required more elaborate functionality for viewing and checking language tags.

We first added support for a project default language tag setting.  The default language is added to the labelling annotation when creating new entities.  This saves a lot of clicking and typing when creating new \interests.

\begin{figure}[t]
    \centering
    \includegraphics[width=0.7\textwidth]{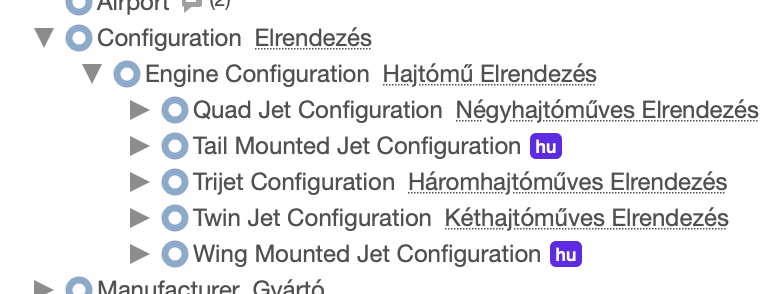}
    \caption{\textbf{An example of secondary language display names}.  The primary display language is ``en'' (English).  The secondary display language is ``hu'' (Hungarian).  The secondary display name is shown to the right of the primary display name.  The colorful tag ``hu'' highlights classes that are missing the ``hu'' language tag.}
    \label{fig:langs}
\end{figure}

We made significant changes to the rendering mechanism in \webprotege so that it is possible to specify a list of \emph{primary and secondary display languages}.  Secondary display languages are used to derive secondary display names for entities, which are displayed along-side the primary display names in the various hierarchies (Figure \ref{fig:langs}) and lists throughout \webprotege.  This provides a context and makes it easier for language specialists to perform translations.

 Finally, we added rule templates, as part of the previously mentioned entity tagging functionality, so that it is possible to display colored indicators next to entities that are missing certain language tags (Figure \ref{fig:langs}).

\section{Production Development}
\label{sec:production}

There were eight \pinterest people involved in the development of the ontology, most of whom are not engineers and did not have any knowledge about OWL ontologies or \webprotege. The eight ontology curators had a one-hour training session on how to use \webprotege, and they were able to start curating content right away. Overall, it took the curators less than one month to build and finalize the ontology. After this first round of curation, some partner teams such as Ads and Sales reviewed the ontology, provided feedback and suggestions for improvements, and they even edited the ontology themselves with minimum training.

Throughout the entire development process there were $\sim$2,000 comments spanning $\sim$1,000 discussion/issue threads on \interests, within \webprotege. Overall the ontology went through 38,000+ revisions before its current version (where each revision involves potentially multiple axiom changes). The final ontology has $\sim$11,000 classes (\interests), 24 verticals (top-level \interests), and up to 12 levels of depth in certain branches of the hierarchy. It contains $\sim$145,000 axioms, out of which $\sim$25,000 are logical axioms and $\sim$95,000 are annotation axioms.

To consume the production ontology, \pinterest engineers built a Python-based pipeline that processes it and generates relational database tables for existing internal applications and other \pinterest tooling pipelines.

\section{Discussion}
\label{sec:discussion}

Throughout the development of the \taxonomy, the \pinterest and \protege teams encountered various challenges both in terms of modeling choices and tool support for ontology development. We describe some of these challenges below, and discuss how they will shape some future directions of our work.

\paragraph{Multiple Inheritance and Cross-Vertical Interests} Some \interests would be best described with multiple parents. We frequently wanted to classify \interests not only by their ``primary type'', but also by their intended role, their location, and the material that they are made from. For example, an \interest in ``Bathroom Lighting'' is an interest in ``Bathroom Decor''. However, ``Bathroom Lighting'' could also be under ``Light Fixtures''.  Allowing multiple parents for ``Bathroom Lighting'' would let advertisers of both bathroom design and lighting stores target \pinners who are interested in ``Bathroom Lighting''.

\paragraph{Multiple Relationship Types among Interests} 
Currently we only have one type of relationship between \interest classes: \textbf{is-a}.  For example, an interest in ``Sandals'' is an interest in ``Shoes''. However, we see a need for more relationships, for example, ``Thanksgiving Recipe'' (a ``Food and Drinks'' \interest) and ``Thanksgiving Decoration'' (a ``Home Decor'' \interest), plus ``DIY Thanksgiving Card'' (a ``DIY and Crafts'' \interest) are all related to the Thanksgiving \interest, so a \pinner interested in one before Thanksgiving time would highly likely be interested in the other two.

\paragraph{New Global Classes and Association with Interests}
Besides categorizing \pins and users against \interests, we can also categorize them into \attributes. For example, ``Colors'', ``Brands'' and ``Materials''. Since \pinterest powers shopping as well, we can imagine understanding supply and demand from the \attribute perspective would give \pinners more relevant products. \attributes should be defined as global (cross-vertical) classes, and they should be associated with applicable \interests via appropriate relationships.

\paragraph{Richer Axiomatization} The ontology has gone through over 38,000 revisions in \webprotege, each composed of potentially multiple axiom changes. The logical axioms used in the ontology are \term{SubClassOf} axioms. The current logical expressivity of the \taxonomy falls under all three OWL 2 profiles: EL, RL, and QL. These profiles benefit from desirable computational properties, such as polynomial time (or less) worst-case complexity for core reasoning tasks. A next iteration of the \taxonomy will make use of existential restrictions (i.e., \term{SomeValuesFrom} class expressions) to support faceted-based classification of \interests. Such an extension would likely not go beyond the expressivity of OWL 2 EL, and would provide the benefit of automatically classifying \interests by their asserted features, such as color and the materials things are made from, whilst preserving a primary axis of asserted classification that is required by some of the downstream consumers of the ontology.


\section{Summary}
\label{sec:conclusion}

During the past year's collaboration with the \protege team, the \pinterest team have concluded that \webprotege is by far the most suitable tool for developing the \taxonomy. Not only has it proved to be a vast improvement upon the spreadsheet-based taxonomy curation, it has worked better than tools developed in-house. Since adopting \webprotege, the curation and development cycle of the \taxonomy has drastically shortened. With the old spreadsheet-based representation, it could have easily taken six months or longer to get to the same stage as we are today.  The process would have been more error prone and far more tedious. \webprotege facilitated the entire development process and allowed \pinterest to build out the \taxonomy in only two months. 

The flexibility of the OWL ontology representation allows the \taxonomy to be easily expanded to a DAG.  Adding further logical axiomatization to encode facets of \interests offers the possibility of improving downstream search and recommendation applications. Besides the concrete outcome of an ontology, the use of \webprotege has had a notable positive effect on strengthening the engagement of internal teams (from advertising and sales) with the Content Management team.  Previously, it was a struggle to do this with the spreadsheet-based editing environment.

In October 2018, \pinterest released the newly developed \taxonomy for interest-based ads targeting with over 1,500 new \interests to target. The advertisers on \pinterest have overall expressed a positive impression of our work on the taxonomy. Additionally, \pinterest have determined that the new representation of their content has measurably increased revenue gains.

Finally, while there are a number of desirable improvements that we could make to the tooling and to the ontology itself, we hope that our experience and insights of using OWL and \webprotege at \pinterest, to model real data from industry, are useful for the Semantic Web community.

\paragraph{\textbf{Acknowledgements}}
We extend a huge thanks to John Milinovich (prev.\ at \pinterest), who played a pivotal role in establishing the collaboration between \pinterest and the \protege team. We also thank Lance Riedel (\pinterest) and Brian Johnson (prev.\ at \pinterest), who steered the project in its earlier stages. The work described in this paper has been fully supported by \pinterest.  Core \webprotege work is supported by NIH NIGMS Grant GM121724.


\bibliographystyle{splncs04}
\bibliography{references}

\end{document}